\documentclass{article}

\usepackage{arxiv}

\usepackage[utf8]{inputenc} % allow utf-8 input
\usepackage[T1]{fontenc}    % use 8-bit T1 fonts
\usepackage{hyperref}       % hyperlinks
\usepackage{url}            % simple URL typesetting
\usepackage{booktabs}       % professional-quality tables
\usepackage{amsfonts}       % blackboard math symbols
\usepackage{nicefrac}       % compact symbols for 1/2, etc.
\usepackage{microtype}      % microtypography
\usepackage{lipsum}		% Can be removed after putting your text content
\usepackage{graphicx}
\usepackage{natbib}
\usepackage{doi}

\usepackage{amsmath}  % for \operatorname
\usepackage{mathtools}  % for \ceil, \floor

\DeclarePairedDelimiter\floor{\lfloor}{\rfloor}  % [How to write ceil and floor in latex? [duplicate]](https://tex.stackexchange.com/a/118217/254188)
\usepackage{amssymb}  % for \varnothing
\usepackage{algorithm, algpseudocode}  % for algorithm; algpseudocode is for \Comment
\usepackage{subcaption}  % for \begin{subfigure}
\usepackage{adjustbox}  % for \begin{adjustbox}, see [How can I center a too wide table?](https://tex.stackexchange.com/a/39436)
\title{CARSS: Cooperative Attention-guided Reinforcement Subpath Synthesis for Solving Traveling Salesman Problem}

\author{
    Yuchen~Shi \\
	Department of  Mathematical Sciences\\
	University of Chinese Academy of Sciences\\
	Beijing 100049, China \\
	\texttt{shiyuchen20@mails.ucas.ac.cn} \\
	\And
	Congying~Han\thanks{Corresponding author} \\
	Department of  Mathematical Sciences\\
	University of Chinese Academy of Sciences\\
	Beijing 100049, China \\
	\texttt{hancy@ucas.ac.cn} \\
	\AND
    Tiande~Guo \\
	Department of  Mathematical Sciences \\
    University of Chinese Academy of Sciences \\
	Beijing 100049, China \\
	\texttt{tdguo@ucas.ac.cn} \\
}

\date{}

\renewcommand{\headeright}{}
\renewcommand{\shorttitle}{CARSS: Cooperative Attention-guided Reinforcement Subpath Synthesis for Solving Traveling Salesman Problem}

\begin{document}
\maketitle

\begin{abstract}
This paper introduces CARSS (Cooperative Attention-guided Reinforcement Subpath Synthesis), a novel approach to address the Traveling Salesman Problem (TSP) by leveraging cooperative Multi-Agent Reinforcement Learning (MARL). CARSS decomposes the TSP solving process into two distinct yet synergistic steps: "subpath generation" and "subpath merging." In the former, a cooperative MARL framework is employed to iteratively generate subpaths using multiple agents. In the latter, these subpaths are progressively merged to form a complete cycle. The algorithm's primary objective is to enhance efficiency in terms of training memory consumption, testing time, and scalability, through the adoption of a multi-agent divide and conquer paradigm. Notably, attention mechanisms play a pivotal role in feature embedding and parameterization strategies within CARSS. The training of the model is facilitated by the independent REINFORCE algorithm. Empirical experiments reveal CARSS's superiority compared to single-agent alternatives: it demonstrates reduced GPU memory utilization, accommodates training graphs nearly 2.5 times larger, and exhibits the potential for scaling to even more extensive problem sizes. Furthermore, CARSS substantially reduces testing time and optimization gaps by approximately 50\% for TSP instances of up to 1000 vertices, when compared to standard decoding methods.
\end{abstract}

% keywords can be removed
% \keywords{Traveling salesman problem \and Cooperative Multi-agent reinforcement learning \and Attention Mechanisms}

\section{Introduction}

The Traveling Salesman Problem (TSP) stands as one of the quintessential combinatorial optimization challenges, seeking the shortest route to visit a set of cities and return to the origin. Its NP-hard nature has spurred continuous research into developing efficient algorithms capable of tackling real-world instances. Traditional methods, such as exact algorithms based on cutting plane method \citep{Chvtal2009SolutionOA} or dynamic programming \citep{Held1962ADP, Bellman1962DynamicPT}, and heuristic algorithms based on insertion \citep{Rosenkrantz1974ApproximateAF}, local search \citep{Helsgaun2000AnEI} or population \citep{Dorigo1997AntCS}, often struggle with scalability and optimality for larger problem sizes, prompting the exploration of innovative paradigms that transcend the limitations of single-agent approaches.

In recent times, the field of Multi-Agent Reinforcement Learning (MARL) has gained prominence as a promising avenue for tackling intricate optimization problems. Notable examples include Level-Based Foraging \citep{Albrecht2013AGM}, Multi-Agent Particle Environment \citep{Mordatch2017EmergenceOG, Lowe2017MultiAgentAF}, StarCraft Multi-Agent Challenge \citep{Samvelyan2019TheSM}, Multi-Robot Warehouse \citep{Christianos2020SharedEA, Dhamankar2020BenchmarkingMD}, Google Research Football \citep{Kurach2020GoogleRF}, and Hanabi \citep{Bard2020TheHC}. Through harnessing the collaborative proficiencies of multiple agents, cooperative MARL brings about the potential to enhance the efficiency of problem-solving processes, overcome computational bottlenecks, and advance scalability. Within this context, we introduce a pioneering algorithm—Cooperative Attention-guided Reinforcement Subpath Synthesis (CARSS)—crafted to transform the approach to solving the Traveling Salesperson Problem (TSP).

CARSS adopts a distinctive two-step strategy to decompose the TSP solving process. The first step, termed "subpath generation", harnesses the power of cooperative MARL to iteratively generate subpaths. Each agent contributes to constructing a subpath, collectively working towards achieving an optimal solution. The second step, "subpath merging," involves the incremental fusion of these subpaths to ultimately form a complete cycle that represents the solution to the TSP. This decomposition not only capitalizes on the strengths of MARL but also strategically divides the problem to mitigate the computational and memory burdens associated with large-scale instances.

A notable feature of CARSS lies in its incorporation of attention mechanisms, which serve a dual role in both feature embedding and parameterization strategies. These mechanisms enhance the agents' ability to capture relevant information and learn effectively from their interactions with the environment. The training of the CARSS model is facilitated by the independent REINFORCE algorithm, a proven reinforcement learning technique.

Our contributions are threefold:

\begin{itemize}
    \item A novel algorithm, CARSS, is introduced for solving the TSP by leveraging cooperative MARL and attention mechanisms. The algorithm decomposes the problem into "subpath generation" and "subpath merging" steps, addressing memory consumption and scalability challenges.
    \item The proposed approach demonstrates substantial improvements in terms of memory efficiency and testing times when compared to conventional single-agent algorithms. CARSS extends the capability to train on larger problem instances while maintaining solution quality.
    \item Empirical results show that the CARSS algorithm reduces testing times and optimization gaps by approximately 50\% for TSP instances of up to 1000 vertices, underscoring its potential to significantly enhance the efficiency of TSP-solving techniques.
\end{itemize}

\section{Related Works}

A considerable portion of the research in the realm of solving the TSP through supervised and reinforcement learning has been rooted in constructive modeling methodologies \citep{Vinyals2015PointerN, Bello2016NeuralCO, Khalil2017LearningCO, Kool2018AttentionLT, Bresson2021TheTN}. These approaches involve the stepwise selection of individual points, akin to methods driven by a singular agent. However, it is noteworthy that these methodologies tend to exhibit elevated time and space complexities when confronted with the task of addressing expansive problem scales. As a testament to this, numerous algorithms demonstrate their efficacy solely on problems of modest proportions, typically up to a size of 200, utilizing a prescribed quantum of GPU resources. For instance, \citet{Joshi2020LearningTR} expound upon the challenges by affirming that "Training on large TSP200 from scratch is intractable and sample inefficient." This intrinsic computational burden consequently restricts their performance when applied to more substantial problem instances. Nevertheless, in contrast to these conventional paradigms, the CARSS algorithm introduces a pioneering approach that strategically decomposes the TSP-solving process into two distinct stages: subpath generation and subpath merging. By leveraging the principles of MARL, CARSS endeavors to surmount the limitations of memory consumption during training, mitigate testing duration, and amplify its scalability.

In the realm of TSP variations, \citet{Zhang2020MultivehicleRP} introduced a MARL-oriented framework addressing the vehicle routing problem encompassing soft time windows for a multi-vehicle scenario. This approach hinged upon predefined regulations, dictating a rotational decision-making process among vehicles. Notably, all vehicles shared a singular policy network, inadvertently rendering the framework functionally akin to single-agent control. Building upon this premise, \citet{Zong2022MAPDPCM} advanced the paradigm by fashioning independent policy networks, eschewing the necessity for predetermined coordination rules in scenarios involving vehicle interaction. This liberation substantially expanded the exploration capacity within the collective of vehicle agents, efficiently tackling the intricacies posed by pickup and delivery problems. Extending this innovation, the CARSS algorithm extrapolates the concept into the domain of TSP, orchestrating a divide-and-conquer methodology tailored to surmount the challenges of larger-scale problem instances. 

\section{Method}

In this section, we present the methodology of CARSS. The subsections that follow outline the cooperative Markov game formulation, algorithm specifics, policy parameterization, policy optimization, and complexity analysis.

\subsection{Cooperative Markov Game Formulation for Traveling Salesman Problem}\label{carss-markov}

Define TSP as the tuple $(\mathcal{I}, f, \mu)$, where $\mathcal{I}$ represents the set of graph instances, $f(G)$ denotes the set of all feasible solutions for the graph $G = (V(G), E(G), w(G))$ within the context of $\mathcal{I}$. Here, each graph $G$ comprises $v(G)$ vertices and $e(G)$ edges. The function $\mu(G, H)$ quantifies the value of solution $H$ within the set $f(G)$ concerning the problem's objective. In the context of TSP, $\mu(G, H)$ equates to $\sum_{e \in H} w_e$, where $w_e$ signifies the weight of edge $e$. The ultimate objective of the problem is to determine the solution $H$ that minimizes this objective value across all instances $G \in \mathcal{I}$, formally expressed as $\operatorname{argmin}_{H \in f(G)} \mu(G, H)$.

For a multi-agent system involving $K \leq \frac{v(G)}{2}$ agents, we can establish the corresponding cooperative Markov game $(K, \mathcal{S}, {\mathcal{A}^k}_{k\in{1,\ldots, K}}, P, r)$ as follows:

\begin{itemize}
	\item The state space, denoted as $\mathcal{S}=\{s \mid s \subseteq H, H \in f(G), G \in \mathcal{I}\}$, encompasses all possible states. The initial state, $s_0$, is represented by the null graph $K_0$, while the state space at the final time step $T$ is $\mathcal{T}=\{H \mid H \in f(G), G \in \mathcal{I}\}$. Each agent shares an identical state at every time step and enjoys full access to all environmental observations.
	\item The action space for agent $k$, noted as $\mathcal{A}^{k}=V(G) \cup E(G)$, includes all vertices and edges of graph $G$. Furthermore, $\mathcal{A}_s^k=(V(G) \setminus V(s)) \cup (E(G) \setminus E(s))$ characterizes the set of feasible actions for agent $k$ within state $s$.
	\item The state transition probability function, $P:\mathcal{S} \times \mathcal{A}^1 \times \cdots \times \mathcal{A}^K \rightarrow \Delta\mathcal{S}$, is defined as follows:
	\begin{equation*}
		P(s^\prime \mid s,a^1,\ldots,a^K) =
		\begin{cases}
		1 & \text{if}\ s \in \mathcal{T} \text{ and } s^\prime = s,\\\\
		1 & \text{if}\ s \in \mathcal{S} \setminus \mathcal{T} \text{ and } s^\prime = s + \sum_{k=1}^{K} a^k,\\\
		0 & \text{otherwise}.
		\end{cases}
	\end{equation*}
	where $\Delta\mathcal{S}$ represents the probabilistic simplex in the state space $\mathcal{S}$, and $s + a$ signifies the disjoint union of graph $s$ and graph $a$.
	\item  The reward function $r:\mathcal{S} \times \mathcal{A}^1 \times \cdots \times \mathcal{A}^K \times \mathcal{S}\rightarrow\delta S$ is defined by $r(s, a^1, \ldots, a^K, s^\prime) = -\mu(G, s^\prime)$. It evaluates to $0$ if $s$ does not belong to $\mathcal{T}$ but $s^\prime$ does; otherwise, it is $0$.
\end{itemize}

Solving TSP involves the acquisition of a strategy denoted as $\pi_{\boldsymbol{\theta}}:\mathcal{S}\rightarrow\Delta(\mathcal{A}^1\times\cdots\times\mathcal{A}^K)$, which is crafted to optimize the expected partial return $J(\boldsymbol{\theta})=\mathbb{E}_{\pi{\boldsymbol{\theta}}}[R_T]$.

In the scenario where the number of agents, denoted as $K=1$, and the approach employed is deep reinforcement learning for solving TSP, the model adheres to the classical reinforcement learning methodology for addressing the TSP \citep{Kool2018AttentionLT, Bresson2021TheTN}. However, this conventional approach exhibits several limitations:

\begin{itemize}
    \item As a TSP route comprises a composition of $v(G)$ edges, invoking the policy network a minimum of $v(G)$ times diminishes the potential benefits of parallel computation within sequential models. Additionally, the substantial computational overhead hampers the training of the model for larger-scale problems.
    \item Policy networks conventionally rely on the computation of the attention matrix, entailing a time and space complexity of $O(v(G)^2)$ \citep{Vaswani2017AttentionIA}. This propensity for excessive memory utilization renders training infeasible for more extensive problem instances.
    \item The actions generated through policy network sampling do not invariably constitute feasible solutions. Consequently, decoding necessitates the application of a mask to regulate the selection of visited vertices. However, as the termination point approaches, the number of visited vertices increases, resulting in a diminished space of viable actions. Consequently, the efficiency of attention matrix computation diminishes, leading to inefficient resource utilization.
\end{itemize}

On the other hand, in scenarios where the number of agents $K>1$, the policy network is invoked a minimum of $v(G)/K$ times. By predetermining the feasible actions for each agent, the number of viable actions per agent is averaged to $v(G)/K$, thereby reducing the time and space complexity of attention matrix computation to $O(\frac{v(G)^2}{K^2})$. This approach also indirectly enhances the efficiency of computational resource utilization, alleviating some of the constraints faced in the single-agent setting.

\subsection{CARSS Algorithm}\label{carss-section}

To address the limitations inherent in solving TSP with a single agent, we propose the CARSS algorithm. Designed for tackling the TSP within Euclidean space, CARSS effectively mitigates these limitations by strategically reducing the action and state space of the underlying Markov game, thereby approximating its optimal strategy.

The CARSS algorithm is structured around two pivotal phases: subpath generation and subpath merging. Within this context, "subpaths" represent non-circular graphs that form integral parts of the problem's final solution tour.

During the subpath generation phase, CARSS initialization features an empty graph, denoted as $K_0$. Each agent independently selects a vertex to ensure non-overlapping choices. Subsequently, during each time step, every agent gradually extends an edge to their selected vertex. This synchronized edge addition results in the simultaneous incorporation of $K$ edges. This process continues until several subpaths of uniform lengths, devoid of intersections, are established. Here, "intersecting" indicates the absence of any intersection between the vertex sets of two subpaths. It is noteworthy that this phase constitutes the majority of the algorithm's runtime due to its computationally intensive nature.

The subsequent subpath mergings phase can be analogously conceived as a single-agent approach. Within this phase, the algorithm connects $K$ subpaths and, at most, $K$ isolated points. This gradual connection is achieved by adding up to $2K$ edges, ultimately culminating in a complete cycle. This phase is crucial for addressing a specific TSP instance of a size not exceeding $4K$. The computation time associated with this phase is nearly negligible due to the relatively diminutive size of the subproblem.

Subsequently, we provide the temporal range encompassing time step $t\in\{1,\ldots,T\}$ within the two phases of subpath generation and subpath merging. Furthermore, we expound upon the precise structure of the space of feasible actions $\mathcal{A}_s^k$ within state $s$ during these two phases.

\subsubsection{Subpath Generation}

In the subpath generation phase, we consider time steps denoted by $t\in\{1,\ldots,T'\}$, where
\begin{equation*}
    T'=
    \begin{cases}
    \frac{v(G)}{K}-2 &\text{if}\ K\text{ evenly divides }v(G),\\\\\
    \floor*{\frac{v(G)}{K}}-1 &\text{otherwise}.
    \end{cases}
\end{equation*}
The rationale for treating the case of $K$ evenly dividing $v(G)$ separately arises from the following consideration: when the algorithm advances to the $(v(G)/K-2)$th time step, a total of $K$ paths exist within the current state, each with a length of $v(G)/K-2$. At this point, the number of visited vertices is $v(G)-K$, and the number of isolated points is $K$. Consequently, by considering isolated points as subpaths with a length of 0, the total count of subpaths to be connected amounts to $2K$. When each agent carries out an additional action, the count of isolated points decreases to $0$, resulting in a reduction of the number of subpaths to $K$. It is evident that a scarcity of isolated points will lead to fewer optional vertices in later stages of the subpath generation phase, thereby reducing training efficiency. Hence, a balance is struck between the number of time steps in the subpath generation phase and the problem size in the subpath merging phase. This strategic compromise enhances overall algorithm performance by marginally decreasing the number of time steps in the subpath generation phase while augmenting the problem's complexity during the subpath merging phase.

At the initial time step $t=1$, the feasible action space for state $K_0$ is defined as $\mathcal{A}_{K_0}^k=V(G)$, where each initial action corresponds to the overlapping initial endpoints of a subpath. For each agent $k$, we respectively denote these front and rear endpoints at its current state as $\operatorname{f}^k(s)$ and $\operatorname{r}^k(s)$. 
For subsequent time steps, when $s\ne K_0$, i.e., when $t\in\{2,\ldots,T'\}$, $\mathcal{A}_s^k$ is determined by addressing the following specialized assignment problem:

\begin{equation*}
\begin{aligned}
\operatorname{minimize} & \sum_{i=1}^{v(G)}\sum_{k=1}^K x_{i,k}\min\{w_{i\operatorname{f}^k(s)},w_{i\operatorname{r}^k(s)}\} & \\
\text{subject to} & \sum_{k=1}^K x_{ik}=1, &\quad i=1,\ldots,v(G) \\
& \sum_{n=1}^{v(G)} \left(x_{i\operatorname{f}^k(s)}+x_{i\operatorname{r}^k(s)}\right)\ge1, &\quad k=1,\ldots,K\\
& x_{ik} \in\{0,1\}, &\quad i=1,\ldots,v(G),\ k=1,\ldots,K
\end{aligned}
\end{equation*}

Here, $x_{ik}$ signifies whether the $i$th vertex is assigned to the $k$th agent. $\min\{w_{i\operatorname{f}^k(s)},w_{i\operatorname{r}^k(s)}\}$ represents the minimum distance from the $i$th vertex to either the first or last endpoint of the path corresponding to the $k$th agent. The objective function aims to minimize the total sum of distances from each vertex to its closest endpoint within the assigned agent's path. The first constraint mandates that each vertex must be assigned to exactly one agent. The second constraint ensures that each agent is assigned at least one vertex, thereby guaranteeing that the length of subpaths generated by each agent consistently increases over time. It's worth noting that the vertex assignment obtained from solving this problem might not necessarily lead to the optimal solution of the original problem. However, it can significantly reduce the action space for the agents, resulting in a substantial acceleration of subpath generation.

A heuristic is designed to efficiently solve the assignment problem. It involves iterating over each agent and having them select the nearest unassigned vertex to fulfill the first constraint. Subsequently, each unassigned vertex is assigned to its nearest agent. The "distances" between vertices $i$ and agents $k$ are defined with respect to the metric $\min_i\{w_{i\operatorname{f}^k(s)},w_{i\operatorname{r}^k(s)}\}$. The complete algorithm for solving this assignment problem is presented in Algorithm \ref{alg:assign}.

Having obtained an approximate solution to the problem, the feasible action space for each agent $k$ in state $s$ is characterized as $\mathcal{A}_s^k=\{(i,j)\in E\mid x_{ik}=1,i\notin V(s), j=\operatorname{argmin}_{j\in\{\operatorname{f}^k(s),\operatorname{r}^k(s)\}}w_{ij}\}$, which denotes an edge in $E$ with unvisited vertices at one end and front and rear vertices of the path corresponding to agent $k$ at the other end. Notably, these sets do not overlap with each other, i.e., $\{i\mid(i,j)\in\mathcal{A}_s^{k_1}\}\cap\{i\mid(i,j)\in\mathcal{A}_s^{k_2}\}=\varnothing,\forall k_1,k_2\in\{1,\ldots,K\},k_1\neq k_2$. This ensures that the states at each time step consist of $K$ disjoint paths.

To enhance the model's viability in addressing large-scale problems, the feasible action space $\mathcal{A}_s^k$ for each agent in this phase is restricted to a maximum of $v(G)/K$ actions, encompassing those closest to the respective agent.

\begin{algorithm}[!htbp]
  \caption{Vertex-agent assignment heurisitic algorithm}\label{alg:assign}
  \begin{algorithmic}[1]
    \Require{Vertex set $V$, Number of agents $K$, Current state $s$}
    \State{Initialize decision variables $x_{ik}\leftarrow 0,\ i=1,\ldots,v(G),\ k=1,\ldots,K$}
    \State{Initialize unassigned vertex list $U\leftarrow V$}
    \For{$k=1,\ldots,K$}
        \State{$i\leftarrow \operatorname{argmin}\{\min\{x_{i\operatorname{f}^k(s)},x_{i\operatorname{r}^k(s)}\}\mid i\in U\}$}
        \State{$x_{ik}\leftarrow 1$}
        \State{$U\leftarrow U\setminus \{i\}$}
    \EndFor
    \While{$U\ne \varnothing$}
        \State{$i\leftarrow U[0]$}
        \State{$k\leftarrow \operatorname{argmin}\{\min\{x_{i\operatorname{f}^k(s)},x_{i\operatorname{r}^k(s)}\}\mid k\in \{1,\ldots,K\}\}$}
        \State{$x_{ik}\leftarrow 1$}
        \State{$U\leftarrow U\setminus \{i\}$}
    \EndWhile
    \State\Return{$\{x_{ij}\}_{i=1,\ldots,v(G),\ k=1,\ldots,K}$}  % [Algorithm return statement does not begin on new line](https://tex.stackexchange.com/a/409443/254188)
  \end{algorithmic}
\end{algorithm}  % [Algorithms](https://www.overleaf.com/learn/latex/Algorithms)

\subsubsection{Subpath Merging}

The subpath merging stage can be conceptualized as addressing a specific variant of TSP using single-agent reinforcement learning. Upon completing the subpath generation, the state $S_{T'}$ encompasses two distinct components: firstly, $K$ disjoint paths each of length $T'$, and secondly, isolated vertices $\{I_i\}_{i\in1,\ldots,v(G)-K(T'+1)}$, which can also be envisaged as $|I|$ paths of length $0$, creating a total of $K+|I|$ paths. To connect these paths into a complete tour, $K+|I|$ extra edges need to be incorporated. This gives rise to a graph $G'$ of size $2(K+|I|)$ constructed as follows:

\begin{align*}
    V(G')&=\{\operatorname{f}^1(S_{T'}),\ldots,\operatorname{f}^K(S_{T'}),I^{\operatorname{f}}_1,\ldots,I^{\operatorname{f}}_{|I|},\operatorname{r}^1(S_{T'}),\ldots,\operatorname{r}^K(S_{T'}),I^{\operatorname{r}}_1,\ldots,I^{\operatorname{r}}_{|I|}\},\\
    E(G')&=\{(i,j)\mid i,j\in V(G'),i\ne j\}\setminus\{(f^1(S_{T'}),r^1(S_{T'})),\ldots,(f^K(S_{T'}),r^K(S_{T'})),\ldots,(I_1^f,I_1^r),\ldots,(I_{|I|}^f,I_{|I|}^r)\}\\
    w(G')&=w(G).
\end{align*}

Here, $I_i^{\operatorname{f}}=I_i^{\operatorname{r}}=I_i$ for $i\in\{1,\ldots,v(G)-K(T'+1)\}$. $V(G')$ consists of the front and rear vertices of each path in state $S_{T'}$. $E(G')$ comprises all potential edges that may be required to amalgamate subpaths into cycles. The edge weights correspond to those in the original graph $G$. Our objective is to ascertain an algorithm that is both highly effective and efficient in order to identify $K+|I|$ edges within the graph $G'$, forming a tour encompassing edges $\{(i,j)\mid|i-j|=K+|I|\}$. This involves selecting an initial vertex and subsequently adding $K+|I|$ edges, thereby rendering $T=T'+K+|I|$. Given that reinforcement learning enables rapid approximate solutions in batch compared to exact algorithms, we employ single-agent reinforcement learning to address this sub-problem. The superscript $1$ indicating the single agent is omitted below for simplicity. The initial vertex $q_{T'}$ can be selected at random from the set of all vertices in $V(G')$. For time steps $t \in \{T'+1, \ldots, T'+K+|I|-1\}$, we begin by determining the other end of the current subpath from vertex $q_{t-1}$, denoted as $p_t$. Subsequently, we derive the set of feasible actions $\mathcal{A}_s \leftarrow \{(p_t, q_t) \mid (p_t, q_t) \in E(G'), j \notin \{q_{T'}, p_{T'+1}, q_{T'+1}, \ldots, p_{t-1}, q_{t-1}\}\}$ and the action chosen at that particular time step is denoted as $(p_t, q_t)$. In the final time step $t = T'+K+|I|$, a terminal edge must be selected to seamlessly integrate the path into a complete cycle.

\begin{figure}[!htbp]
    \includegraphics[width=\textwidth, angle=0]{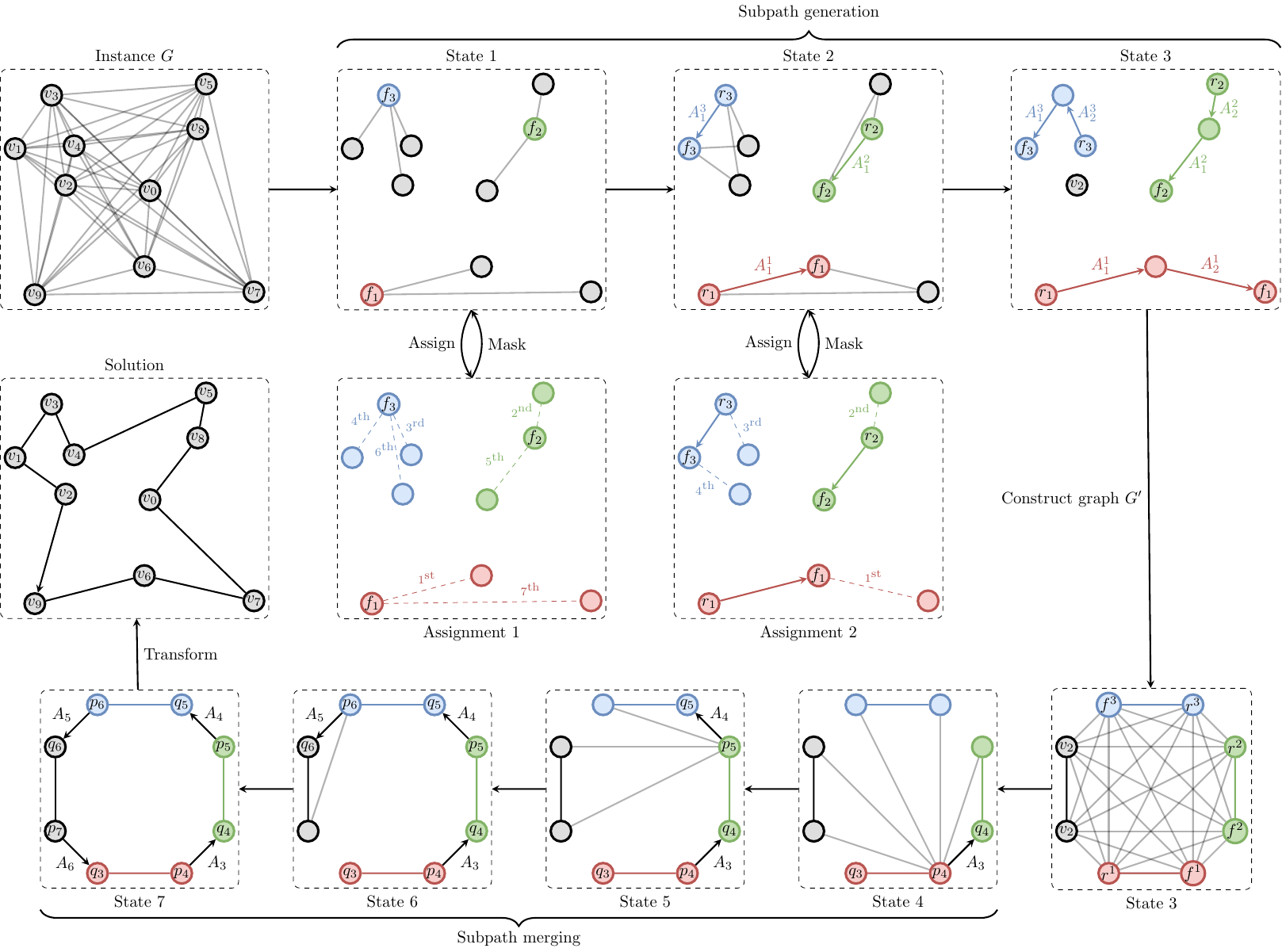}
	\caption{Illustration of CARSS Algorithm for TSP Solving. In this instance, there are 10 vertices, with the number of agents set to $K=3$, termination time of subpath generation set to $T'=2$, and the number of isolated vertices denoted as $|I|=1$. The solid gray lines and arrowed lines represent the action space and the selected actions, respectively. This example entails solving assignment subproblems during two rounds of subroute generation. Within each assignment, dashed lines illustrate the assignment relationships between agent subpaths and unconnected vertices, with corresponding labels indicating the order of assignments.}
	\label{fig:carss-example}
\end{figure}

The comprehensive CARSS algorithm is depicted in Algorithm \ref{alg:carss}, while an illustrative example is showcased in Figure \ref{fig:carss-example}.

\begin{algorithm}[!htbp]
  \caption{Cooperative Attention-guided Reinforcement Subpath Synthesis (CARSS) Algorithm}\label{alg:carss}
  \begin{algorithmic}[1]
    \Require{Graph $G=(V,E,w)$, number of agents $K$, starting vertices of agents $v_1,\ldots,v_K\in V$}
    \State{$s\leftarrow G[\{v_1,\ldots,v_K\}]$}
    \For{$k=1,\ldots,K$}
        \State{$\operatorname{f}^k(s)\leftarrow v_k$}
        \State{$\operatorname{r}^k(s)\leftarrow v_k$}
    \EndFor
    \State{$T'\leftarrow\floor*{\frac{v(G)}{K}}-(1\text{ if }K\text{ divides }v(G)\text{ else }0)$}
    \For{$t=1,\ldots,T'$}
        \State{Apply Algorithm \ref{alg:assign} with $(V,K,s)$ to obtain $\{x_{ij}\}_{i=1,\ldots,v(G),\ k=1,\ldots,K}$}
        \State{$\mathcal{A}_s^k\leftarrow\{(i,j)\in E\mid x_{ik}=1,i\notin V(s),j=\operatorname{argmin}_{j\in\{\operatorname{f}^k(s),\operatorname{r}^k(s)\}}w_{ij}\},k=1,\ldots,K$}
        \State{Apply parameterized policy to obtain $A^k_t\in\mathcal{A}_s^k,k=1,\ldots,K$}
        \State{$s\leftarrow s+\sum_{k=1}^K A^k_t$}
        \State{Update $\operatorname{f}^k(s)$ and $\operatorname{r}^k(s)$}
    \EndFor
    \State{$I\leftarrow V(G)\setminus V(s)$}\Comment{$|I|=v(G)-K(T'+1)$}
    \State{$\{I^f_1,\ldots,I^f_{|I|}\}\leftarrow I$}
    \State{$\{I^r_1,\ldots,I^r_{|I|}\}\leftarrow I$}
    \State{$V(G')\leftarrow\{\operatorname{f}^1(S_{T'}),\ldots,\operatorname{f}^K(S_{T'}),I^{\operatorname{f}}_1,\ldots,I^{\operatorname{f}}_{|I|},\operatorname{r}^1(S_{T'}),\ldots,\operatorname{r}^K(S_{T'}),I^{\operatorname{r}}_1,\ldots,I^{\operatorname{r}}_{|I|}\}$}
    \State{$E(G')\leftarrow\{(i,j)\mid i,j\in V(G'),i\ne j\}\setminus\{(f^1(S_{T'}),r^1(S_{T'})),\ldots,(f^K(S_{T'}),r^K(S_{T'})),\ldots,(I_1^f,I_1^r),\ldots,(I_{|I|}^f,I_{|I|}^r)\}$}
    \State{$w(G')\leftarrow w(G)$}
    \State{$q_{T'}\leftarrow\operatorname{RandomSelect}(V(G'))$}
    \For{$t=T'+1,\ldots,T'+K+|I|-1$}
        \State{$p_t\leftarrow$ the other end of the subpath of vertex $q_{t-1}$}
        \State{$\mathcal{A}_s\leftarrow\{(p_t,q_t)\mid (p_t,q_t)\in E(G'),j\notin\{q_{T'},p_{T'+1},q_{T'+1},\ldots,p_{t-1},q_{t-1}\}\}$}
        \State{Apply parameterized policy to obtain $(p_t,q_t)=A_{t}\in\mathcal{A}_s$}
        \State{$s\leftarrow s+A_t$}
    \EndFor
    \State{$p_{T'+K+|I|}\leftarrow$ the other end of the subpath of vertex $q_{T'+K+|I|-1}$}
    \State{$A_{T'+K+|I|}\leftarrow(p_{T'+K+|I|},q_{T'})$}
    \State{$s\leftarrow s+A_{T'+K+|I|}$}
    \State\Return{$s$}
  \end{algorithmic}
\end{algorithm}  % [Algorithms](https://www.overleaf.com/learn/latex/Algorithms)

\subsection{Policy Parameterization}

\subsubsection{Subpath Generation Parameterization}

The input data to the subpath generation stage within the CARSS algorithm comprises the 2D Euclidean spatial coordinates of the graph's vertex set $V(G)$, represented as $X\in\mathbb{R}^{v(G)\times 2}$. The action index $U^k_t$ is determined by the probability distribution of feasible actions for agent $k$ at time step $t$ within state $S_t$, denoted as $\pi^d_{\boldsymbol{\theta}}(U^1_t, \ldots, U^K_t \mid G, S_t)$. This index is selected from the set $\{1, \ldots, v(G)/K\}$, ultimately defining the action $A^k_t$.

To efficiently capture information about neighboring vertices for each vertex, we employ a mapping from the 2D vertex coordinates $X\in\mathbb{R}^{v(G)\times 2}$ to higher-dimensional vertex features $h^v_i,i=1,\ldots,v(G)$:

\begin{align*}
    H^v_{l}&=\operatorname{FFN}\left(\operatorname{MHA}\left(H^v_{l-1},H^v_{l-1},H^v_{l-1},H^v_{l-1},J_{v(G)}\right)\right)\in\mathbb{R}^{v(G)\times d_v},\\
    H^v_{L^{\text{enc}_v}}&=[h^v_1,\cdots,h^v_{v(G)}]^T.
\end{align*}

Here, $H^v_0=XW^x+\boldsymbol{1}_{v(G)}b^x$, where $W^x\in \mathbb{R}^{2\times d_v}$ and $b^x\in\mathbb{R}^{1\times d_v}$ are trainable parameters. The parameter $d_v$ indicates the dimension of vertex features, $l\in\{1,\ldots L^{\text{enc}_v}\}$ represents the layer index of the vertex encoder within the subpath generation stage, and $J_{v(G)}$ denotes the $v(G)$-order square matrix filled with $1$ entries.

Drawing inspiration from Zong Zefang et al.'s MAPDP paper \citep{Zong2022MAPDPCM}, we adopt a novel approach by concatenating the feature vectors of both the front and rear vertices across all agents. This strategy facilitates the sharing of positional information among agents and yields the global information feature vector for agent $k$:

\begin{equation*}
    \operatorname{comm}^k_t=\operatorname{FFN}([h^v_{\operatorname{f}^k(S_t)},h^v_{\operatorname{r}^k(S_t)}])\in\mathbb{R}^{1\times d_v}.
\end{equation*}

Subsequently, it is concatenated with the feature vectors of the agent's starting and ending vertices (comprising the comprehensive feature representation of the visited vertices), thus forming the feature vector for agent $k$:

\begin{equation*}
    \operatorname{context}^k_t=\operatorname{FFN}([h^v_{\operatorname{f}^k(S_t)},h^v_{\operatorname{r}^k(S_t)},\frac{1}{|v(G)-v(S_t)|} \sum_{i\in V(G)\setminus V(S_t)}h^v_i),\operatorname{comm}^k_t])\in\mathbb{R}^{1\times d_v}.
\end{equation*}

To facilitate cooperative interactions, a multi-head attention mechanism is then employed on the feature vectors of the aforementioned $K$ agents. This mechanism enhances the exchange of vital information among the agents:

\begin{align*}
    H^a_{t,0}&=[\operatorname{context}^1_t;\ldots; \operatorname{context}^K_t]\in\mathbb{R}^{K\times d_v},\\
    H^a_{t,l}&=\operatorname{FFN}\left(\operatorname{MHA}\left(H^a_{t,l-1},H^a_{t,l-1},H^a_{t,l-1},H^a_{t,l-1},H^a_{t,l-1},J_{K}\right)\right)\in\mathbb{R}^{K \times d_v}.
\end{align*}

Here, $l\in\{1,\ldots L^{\text{enc}_a}\}$ denotes the layer index of the agent encoder, and $J_K$ represents a $K$-order square matrix composed entirely of $1$ entries.

Inspired by the utilization of the original Transformer model by \citet{Bresson2021TheTN} for addressing the TSP, we have introduced a novel concept of memory vectors with increasing lengths over time. These vectors enhance the model's capacity to incorporate historical information progressively:

\begin{equation*}
    \operatorname{memory}_t=[\operatorname{memory}_{t-1};h^v_{i:(i,j)=A_{t-1}}]\in\mathbb{R}^{K\times t\times d_v}.
\end{equation*}

Subsequently, these feature vectors and memory vectors of the agents are harnessed as inputs to the multi-head attention mechanism. This integration enables the model to adeptly capture the characteristics of partial solutions:

\begin{align*}
    h^d_{t,0}&=\operatorname{Reshape}\left(H^a_{t,\text{enc}_a},(K,1,d_v)\right),\\
    h^d_{t,l}&=\operatorname{FFN}(\operatorname{MHA}(h^d_{t,l-1},\operatorname{memory}_t,\operatorname{memory}_t))\in\mathbb{R}^{K\times 1\times d_v}.
\end{align*}

Here, the operation $\operatorname{Reshape}$ alters the dimensions of the tensor $H^a_{t,\text{enc}_a}$ from $\mathbb{R}^{K\times d_v}$ to $\mathbb{R}^{K\times 1\times d_v}$, preserving its elements while aligning its dimensions with those of the memory vector $\operatorname{memory}_t$ at time step $t$. $l\in\{1,\ldots L^{\text{dec}_g}\}$ designates the decoder layer index.

Subsequently, we construct feature vectors for each assigned vertex of every agent along with their corresponding masks. It should be noted that in Subsection \ref{carss-section}, the solution to the assignment problem $x_{ik}$ has been approximated using a heuristic algorithm. Each non-zero element of $x_{ik}$ represents the vertex $i$ being assigned to the agent $k$. However, the number of vertices assigned to each agent, denoted as $\alpha_k = |\{i\mid x_{ik}=1\}|$, varies. To enable an efficient computation of the policy, we propose selecting feature vectors of $v(G)/K$ vertices that are assigned to each agent based on their proximity. If the feasible actions are fewer than $\frac{v(G)}{K}$, we supplement the feature vectors with those of arbitrarily chosen infeasible vertices. This process yields the feature vector $\operatorname{assign}_t$, which represents the assignments. Subsequently, we utilize the mask $M_t^{\operatorname{assign}}$ to prevent the model from generating these additionally introduced actions. This can be formally expressed as follows:

\begin{align*}
    \operatorname{assign}^k_t&=\begin{cases}
        \left[[h^v_i]_{i:x_{ik}=1};[h^v_i]_{i:x_{ik}=0}[:\frac{v(G)}{K}-\alpha_k]\right] &\text{if }\alpha_k<\frac{v(G)}{K},\\
        [h^v_i]_{i:x_{ik}=1}[:\frac{v(G)}{K}] &\text{otherwise}.
    \end{cases}\in\mathbb{R}^{\frac{v(G)}{K}\times d_v},\\
    \operatorname{assign}_t&=[\operatorname{assign}^1_t;\ldots;\operatorname{assign}^K_t]\in\mathbb{R}^{K\times\frac{v(G)}{K}\times d_v },\\
    M_t^{\operatorname{assign}}&=\{M_{tij}^{\operatorname{assign}}\}_{i=1,\ldots,\frac{v(G)}{K},k=1,\ldots,K}=\begin{cases}
        1&\text{if }x_{\operatorname{map}(i)k}=1,\\
        0&\text{otherwise}.
\end{cases}\in\mathbb{R}^{K\times 1\times\frac{v(G)}{K}}.
\end{align*}

Here, the notation $[:i]$ signifies selecting the first $i$ rows of the matrix, and the function $\operatorname{map}(\cdot)$ serves to associate the mask index back to the vertex index of the initial instance $G$ during the subpath generation stage.

Finally, we proceed to identify the index of the selected feasible action at time step $t$ for each agent $k$ within the assignment list, denoted as $U^k_t\in\{1,\ldots,\frac{v( G)}{K}\}$. Subsequently, we map $U^k_t$ back to the original graph's vertex index and determine the adjacent edge closer to the front and rear of that agent. This mapped vertex and edge combination serves as the action $A^k_t$ at time step $t$, with $t$ ranging from $1$ to $T'$. The probability distribution of $U^1_t,\ldots,U^K_t$ is formulated using the input features of the agents, namely $h^d_{t,L^{\text{dec}_g}}$, the vertex feature vectors for assignment denoted as $\operatorname{assign}_t$, and the mask $M^{\operatorname{assign}}_t$. This distribution is computed according to the following equation:

\begin{equation*}
    \pi^d_{\boldsymbol{\theta}}(U^1_t,\ldots,U^K_t\mid G,S_t)=\operatorname{softmax}\Bigg(C\operatorname{tanh}\Big(M^{\operatorname{ assign}}_t\odot(M^{\operatorname{tanh}})\Big)(h^d_{t,L^{\text{dec}_g}}W^d_1+b^d_1)(\operatorname{assign}_t W^d_2+\boldsymbol{1}_{K}b^d_2)^T\slash\sqrt{d}\Big)\Bigg).
\end{equation*}

Here, $C=10$ serves as a cropping threshold. The parameters $W^d_1,W^d_2\in\mathbb{R}^{d_v\times d_v}$ and $b^d_1,b^d_2\in\mathbb{R}^{1\times d_v}$ are trainable parameters.

Once the index $U^k_t$ for agent $k$ within the assignment vector at time step $t$ is determined, the corresponding action can be computed as:

\begin{equation*}
    A^k_t=\left(\operatorname{map}(U^k_t),j=\operatorname{argmin}_{j\in\{\operatorname{f}^k( S_t),\operatorname{r}^k(S_t)\}}w_{\operatorname{map}(U^k_t)j}\right).
\end{equation*}

Here, the operation $\operatorname{map}(\cdot)$ serves to establish a correspondence between the indices within the mask and the vertices of instance graph $G$. This process results in connecting that vertex to the nearest endpoint along the subpath represented by agent $k$.

\subsubsection{Subpath Merging Parameterization}

The input to the subpath merging phase in the CARSS algorithm is represented by the graph denoted as $G^{\prime}$. This graph is utilized in conjunction with the probability distribution of feasible actions at time $t$ under the state $S_t$, denoted as $\pi^c_{\boldsymbol{\theta}}(U_t\mid G^{\prime},S_t)$. The selection of the vertex index $U_t$ from the graph $G^{\prime}$ is a crucial step in determining the eventual action $A_t$, with $U_t$ taking values from the set $\{1,\ldots,2(K+|I|)\}$.

It's important to note that the input graph size for this phase is $2(K+|I|)$, where the vertices comprise the front and rear endpoints of the paths corresponding to each agent from the previous phase, and the total number of edges to be added is $K+|I|$. In order to facilitate the efficient extraction of information from the opposite end of the road as well as from other vertices within the neighborhood, a two-step process is employed. Initially, the two-dimensional vertex coordinates of the front (rear) vertex are concatenated with the two-dimensional coordinates of the rear (front) vertex. Subsequently, this amalgamated information is projected into a higher-dimensional vertex feature space denoted as ${h'}^{v}_{i},i=1,\ldots,v(G')$. This transformation can be expressed as follows:

\begin{align*}
    X'&=[[X_{\operatorname{f}^1(S_{T'})},X_{\operatorname{r}^1(S_{T'})}];\ldots;[X_{\operatorname{f}^K(S_{T'})},X_{\operatorname{r}^K(S_{T'})}];[X_{I_1},X_{I_1}];\ldots;[X_{I_{v(G)-KT'}},X_{I_{v(G)-KT'}}]]\in\mathbb{R}^{v(G')\times 4}\\
    H^{v'}_0&=X'W^{x'}+\boldsymbol{1}_{v(G)}b^{x'}\in\mathbb{R}^{v(G')\times d_v}\\
    H^{v'}_{l}&=\operatorname{FFN}\left(\operatorname{MHA}\left(H^{v'}_{l-1},H^{v'}_{l-1},H^{v'}_{l-1},H^{v'}_{l-1} ,J_{v(G')}\right)\right)\in\mathbb{R}^{v(G')\times d_v},\\
    H^{v'}_{L^{\text{enc}_{v'}}}&=[h^{v'}_1,\cdots,h^{v'}_{v(G')}]^T.
\end{align*}

Here, $X_{i}\in\mathbb{R}^{1\times 2}$ signifies the coordinates of the $i$th vertex in graph $G$. The parameters $W^{x'}\in \mathbb{R}^{4\times d_v}$ and $b^{x'}\in\mathbb{R}^{1\times d_v}$ are trainable parameters. The dimensionality of the feature vector is denoted by $d_v$. The layer index of the vertex encoder at the subpath merging stage is denoted by $l\in\{1,\ldots L^{\text{enc}_{v'}}\}$, and $J_{v(G)}$ represents a square matrix of order $v(G)$ with all elements equal to $1$.

Building upon Kool et al.'s pioneering work in employing reinforcement learning for solving pathfinding problems \citep{Kool2018AttentionLT}, we adopt a similar approach to formulate the feature representation of states, allowing us to effectively capture the relevant information from the graph's vertex features and incorporate it into the state representation. Specifically, we construct the feature representation by averaging the vertex feature vectors on the graph, combining the feature vector of the initially selected vertex, and concatenating it with the feature vector of the previously chosen vertex in the sequence of steps"

\begin{align*}
    h_{\operatorname{graph}} &= \frac{1}{v(G^{\prime})}\sum_{i=1}^{v(G^{\prime})}h^{v^{\prime}}_i\in\mathbb{R}^{1\times d_v}\\
    h_{\operatorname{front}} &= h^{v^{\prime}}_{U_{T^{\prime}+1}}\in\mathbb{R}^{1\times d_v}\\
    h_{\operatorname{rear}} &= h^{v^{\prime}}_{U_{t}}\in\mathbb{R}^{1\times d_v}\\
    h_{\operatorname{state}} &= [h_{\operatorname{graph}},h_{\operatorname{front}},h_{\operatorname{rear}}]W^s+\boldsymbol{1}_{v(G^{\prime})}b^{s}\in\mathbb{R}^{v(G^{\prime})\times d_v}
\end{align*}

Here, $W^s \in \mathbb{R}^{3d_v \times d_v}$ and $b^s \in \mathbb{R}^{1 \times d_v}$ are trainable parameters. The parameter $U_{t}$ corresponds to the vertex indices selected from the graph $G'$ during the subpath merging phase at time step $t$.

Ultimately, we determine the index $U_t\in\{1,\ldots,K+|I|\}$ of the viable action chosen at time $t$ from the set of vertices in graph $G'$ using the parameterized strategy $\pi^c_{\boldsymbol{\phi}}(U_t, G,S_t)$. This index is then translated back to the vertex indexes of the original graph to yield action $A_t$ at time $t$, where $t\in\{T'+1,\ldots,T'+K+|I|-1\}$. The probability distribution of $U_t$ is computed based on the input feature vector of states $h^{p'}_{t,0}=h^{\operatorname{state}}$, the vertex feature vector of graph $G'$, denoted as $H^{v'}_{L^{\text{enc}_{v'}}}$, and the mask $M_t$. The calculation follows this pattern:

\begin{align*}
    h^{p'}_{t,l}&=\operatorname{MHA}(h^{p'}_{t,l-1}, H^{v'}_{L^{\text{enc}_{v'}}}, H^{v'}_{L^{\text{enc}_{v'}}}, M_t),\\
    \pi^c_{\boldsymbol{\phi}}(U_t\mid G,S_t)&=\operatorname{softmax}\Bigg(C\operatorname{tanh}\Big(M_t\odot(h^{p'}_{t,L^{\text{dec}_{c}}}W^c_1+b^c_1)(H^{v'}_{L^{\text{enc}_{v'}}}W^c_2+\boldsymbol{1}_{v(G')}b^c_2)^T\slash\sqrt{d}\Big)\Bigg).
\end{align*}

In the equation above, $l\in\{1,\ldots L^{\text{dec}_{c}}\}$ denotes the decoder layer index specific to the subpath merging stage. The parameter $C$ is set to $10$ as a cropping threshold, while $M_t$ represents the vertex mask that has yet to be accessed. $W^c_1,W^c_2\in\mathbb{R}^{d_v\times d_v}$ and $b^c_1,b^c_2\in\mathbb{R}^{1\times d_v}$ are trainable parameters. Once the vertex index $U_t$ is established on graph $G'$, chosen by agent $k$ at time $t$, the corresponding action becomes

\begin{equation*}
    A_t=\left(V_{U_{t-1}+(K+|I|)\cdot(\operatorname{bool}(U_{t-1}<K+|I|))}(G '),V_{U_t}(G')\right).
\end{equation*}

Here, $V_i(G')$ denotes the $i$-th vertex in graph $G'$, and $\operatorname{bool}(\cdot)$ functions as a logic operator returning either $0$ or $1$. It serves to determine whether $U_{t-1}$ corresponds to the front or rear vertex. If it represents the front vertex, its index is increased by $K+|I|$ to ensure that the newly selected vertex at time $t$ connects to its endpoint. Conversely, if it represents the rear vertex, its index is decreased by $K+|I|$ to link it properly.

At the final time step $t=T'+K+|I|$, a decisive selection of a singular edge ensures the formation of a cycle. As a result, the need for parameterized strategies is obviated.

\subsection{Policy Optimization}

This section introduces optimization methods for the parameterized strategies involved in the CARSS algorithm's subroute generation and subroute merging phases. In the subroute generation phase, initially, a set of $N$ groups of starting vertices $\{U^{n,1}_{0},\ldots,U^{n,K}_{0}\}_{n=1}^N$ is selected within the vertex set $V(G)$, ensuring distinct vertices within each group. Subsequently, leveraging the probability distribution of the policy $\pi^{d}_{\boldsymbol{\theta}}$, $N$ trajectories are sampled for the same instance. This results in a sequence of states, assignment vector indices, and reward trajectories $\{(S_{t-1}^n,U_{t-1}^{n,1},\ldots,U_{t-1}^{n,K},R_t^n)_{t=1}^{T^{\prime}}\}_{n=1}^N$, where $K$ is the number of agents and $T'$ is the termination time of the subroute generation phase. 

Moving to the subroute merging phase, a choice is made to establish $2(K+|I|)$ sets of initial vertex indices for time $T^{\prime}$ in graph $G^{\prime}$. Specifically, indices are assigned as $U_{T^{\prime}+1}^{1,1}=U_{T^{\prime}+1}^{2,1}=\ldots=U_{T^{\prime}+1}^{N,1}=1,\ldots,U_{T^{\prime}+1}^{1,2(K+|I|)}=U_{T^{\prime}+1}^{2,2(K+|I|)}=\ldots=U_{T^{\prime}+1}^{N,2(K+|I|)}=2(K+|I|)$. Then, employing the probability distribution of policy $\pi^{c}_{\boldsymbol{\phi}}$, each of the $2(K+|I|)$ indices is sampled independently. This process yields another sequence of states, vertex indices in graph $G^{\prime}$, and reward trajectories $\{\{(S_{t-1}^{n,m},U_{t-1}^{n,m},R_t^{n,m})_{t=T'}^{T'+K+|I|}\}_{m=1}^{2(K+|I|)}\}_{n=1}^{N}$. Here, $I=V(G)\setminus V(S_{T^{\prime}})$ signifies the set of isolated points in graph $G$ at the end of the subroute generation phase, and $V_i(G^{\prime})$ represents the $i$-th vertex in graph $G^{\prime}$.

It is important to note, as defined Section \ref{carss-markov}, that the reward functions are structured such that $R_t^n=0$ for all $t\in\{1,\ldots,T^{\prime}\}$ and $R_t^{n,m}=0$ for all $t\in\{T^{\prime}+1\ldots,T^{\prime}+K+|I|\}$, and for all $n\in\{1,\ldots,N\}$ and $m\in\{1,\ldots,2(K+|I|)\}$. Non-zero rewards are solely associated with $R_{T^{\prime}+K+|I|}^{n,m}$, for all $n\in\{1,\ldots,N\}$ and $m\in\{1,\ldots, 2(K+|I|)\}$, representing the final circuit length.

When considering the policy gradient for each agent, the remaining agents are treated as part of the environment. With reference to the Policy Gradient Theorem \citep{Sutton1999PolicyGM}, the gradient of the expected cumulative reward can be approximated as follows:

\begin{align*}
    \nabla J(\boldsymbol{\theta})&\approx \frac{1}{N}\sum_{n=1}^{N}\frac{1}{K}\sum_{k=1}^{K}\left(\min_{m\in\{1,\ldots,2(K+|I|)\}}R_{T'+K+|I|}^{n,m}-b^d\right)\nabla\log\prod_{t=1}^{T'} \pi_{\boldsymbol{\theta}}(U_t^{n,k}\mid G,S_t^{n,k}),\\
    \nabla J(\boldsymbol{\phi})&\approx\frac{1}{N}\sum_{n=1}^{N}\frac{1}{2(K+|I|)}\sum_{m=1}^{2(K+|I|)}\left(R_{T'+K+|I|}^{n,m}-b^c\right)\nabla\log\prod_{t=T'+1}^{T'+K+|I|}\pi_{\boldsymbol{\phi}}(U_t^{n,m}\mid G',S_t^{n,m}).
	.
\end{align*}

Here, $b^d=\frac{1}{N}\sum_{n=1}^N \min_{m\in\{1,\ldots,2(K+|I|)\}}R_{T'+K+|I|}^{n,m}$ and $b^c=\frac{1}{2(K+|I|)}\sum_{m=1}^{2(K+|I|)}R_{T'+K+|I|}^{n,m}$. The former corresponds to the Policy Optimization with Multiple Optima (POMO) baseline \citep{Kwon2020POMOPO} obtained by sampling the decoding of merged subpaths from the $2(K+|I|)$ randomly selected vertices in the subpath merging stage. The latter signifies the POMO baseline obtained by sampling $N$ vertices decoded from randomly chosen vertices in the subpath generation stage.

We trained the model using the independent REINFORCE algorithm \citep{Williams1992SimpleSG} with the POMO baseline, employing the Adam optimizer \citep{Kingma2015AdamAM} for parameter updates. The training procedure is detailed in Algorithm \ref{alg:carss-reinforce}.

\begin{algorithm}[!htbp]
  \caption{Independent REINFORCE algorithm with Policy Optimization Multiple Optima baseline}\label{alg:carss-reinforce}
  \begin{algorithmic}[1]
    \Require{Number of iterations $E$, batch size $B$, trajectory samples per instance $N$}
    \State{Initialize $\boldsymbol{\theta},\boldsymbol{\phi}$}
    \For{$\text{epoch}=1,\ldots,E$}
      \For{$i=1,\ldots,B$}
        \State{$G_i\leftarrow\operatorname{RandomInstance()}$}
        \State{$T'\leftarrow\floor*{\frac{v(G)}{K}}-(1\text{ if }K\text{ divides }v(G)\text{ else }0)$}
        \For{$n=1,\ldots,N$}
          \For{$k=1,\ldots,K$}
            \State{$U_{0}^{n,k}\leftarrow\operatorname{RandomSelect}\left(V(G_i)\setminus\{U_{0}^{n,k'}\}_{k'\in\{1,\ldots,k-1\}}\right)$}
          \EndFor
          \State{$\{(S_{t-1}^n,U_{t-1}^{n,1},\ldots,U_{t-1}^{n,K},R_t^n)\}_{t=1}^{T'}\leftarrow\operatorname{Rollout}(G_i,\pi_{\boldsymbol{\theta}})$}
          \State{$I\leftarrow V(G_i)\setminus V(S^n_{T'})$}
          \State{$G'_i\leftarrow\operatorname{ConstructSubgraph}(G_i,S^{n}_{T'})$}
          \State{$U_{T'+1}^{n',m'}\leftarrow m,n'\in\{1,\ldots,N\},m'\in\{1,\ldots,2(K+|I|)\}$}
          \State{$\{\{(S_{t-1}^{n,m},U_{t-1}^{n,m},R_t^{n,m})\}_{t=T'+1}^{T'+K+|I|}\}_{m=1}^{2(K+|I|)}\leftarrow\operatorname{Rollout}(G'_i,\pi_{\boldsymbol{\phi}})$}
        \EndFor
        \State{$b^d=\frac{1}{N}\sum_{n=1}^N \min_{m\in\{1,\ldots,2(K+|I|)\}}R_{T'+K+|I|}^{n,m}$}
        \State{$b^c\leftarrow\frac{1}{2(K+|I|)}\sum_{m=1}^{2(K+|I|)}R_{T'+K+|I|}^{n,m}$}
        \State{$\nabla J(\boldsymbol{\theta})\leftarrow\frac{1}{N}\sum_{n=1}^{N}\frac{1}{K}\sum_{k=1}^{K}\left(\min_{m\in\{1,\ldots,2(K+|I|)\}}R_{T'+K+|I|}^{n,m}-b^d\right)\nabla\log\prod_{t=1}^{T'} \pi_{\boldsymbol{\theta}}(U_t^{n,k}\mid G,S_t^{n,k})$}
        \State{$\nabla J(\boldsymbol{\phi})\leftarrow\frac{1}{N}\sum_{n=1}^{N}\frac{1}{2(K+|I|)}\sum_{m=1}^{2(K+|I|)}\left(R_{T'+K+|I|}^{n,m}-b^c\right)\nabla\log\prod_{t=T'+1}^{T'+K+|I|}\pi_{\boldsymbol{\phi}}(U_t^{n,m}\mid G',S_t^{n,m})$}
        \State{$\boldsymbol{\theta}\leftarrow\operatorname{Adam}(\boldsymbol{\theta}, \nabla J(\boldsymbol{\theta}))$}
        \State{$\boldsymbol{\phi}\leftarrow\operatorname{Adam}(\boldsymbol{\phi}, \nabla J(\boldsymbol{\phi}))$}
      \EndFor
    \EndFor
  \end{algorithmic}
\end{algorithm}  % [Algorithm in IEEE format](https://tex.stackexchange.com/a/219820/254188)

\subsection{Complexity Analysis}

In this section, we present an analysis of the overall time and space complexity of the CARSS algorithm. This analysis highlights the advantages of our approach in terms of complexity compared to classical reinforcement learning-based methods for solving TSP \citep{Kool2018AttentionLT, Bresson2021TheTN}. Moreover, it underscores the potential for training on larger problem instances.

Firstly, let's consider the algorithm's time complexity. This algorithm involves the utilization of the self-attention mechanism from multiple Transformer models during both the encoding and decoding processes. For a sequence of length $n$, the time complexity of this algorithm is determined by $O(n^2 d + nd^2)$, as discussed in \citet{Vaswani2017AttentionIA}, where $d$ represents the model's dimension. For the sake of simplicity, we can omit the term $nd^2$, particularly since during algorithm execution, the difference between $n$ and $d$ tends to be marginal or in scenarios where $n>d$. Additionally, we will disregard certain network-specific parameters like $L^{\text{enc}_a}$, $L^{\text{dec}_g}$, etc. Furthermore, our analysis will focus solely on the case where a single trajectory is sampled in both phases ($N=1$). The computational approach is as follows:

\begin{equation*}
    O\left(\underbrace{\left(\overbrace{K^2}^{H^a_{t,L^{\text{enc}_a}}}+ \overbrace{KT^{\prime}}^{h^d_{t,L^{\text{dec}_g}}}+\overbrace{K\frac{v(G)}{K}}^{\pi^d_{\boldsymbol{\theta}}}\right)\cdot T^{\prime}d}_{\text{Subpath generation}}+\underbrace{\overbrace{(K+|I|)}^{\pi^d_{\boldsymbol{\phi}}}\cdot (K+|I|)d}_{\text{Subpath merging}}\right)=O\left(\left(Kv(G)+2\frac{\left(v(G)\right)^2}{K}+4K^2\right)d\right),
\end{equation*}

Next, we delve into the consideration of the algorithm's space complexity. For a sequence of length $n$, the space complexity of the self-attention module is $O(n^2)$, as indicated by \citet{Vaswani2017AttentionIA}. However, recent advancements in the field have demonstrated that for encoders, this complexity can be reduced to $\sqrt{n}$ \citep{Rabe2021SelfattentionDN}. By employing this technique, the space complexity can be expressed as follows:

\begin{equation*}
    O\left(\underbrace{\overbrace{K\sqrt{K}}^{H^a_{t,L^{\text{enc}_a}}}+\overbrace{K\left(T^{\prime}\right)^2}^{h^d_{t,L^{\text{dec}_g}}}+\overbrace{K\left(\frac{v(G)}{K}\right)^2}^{\pi^d_{\boldsymbol{\theta}}}}_{\text{Subpath generation}}+\underbrace{\overbrace{(K+|I|)^2}^{\pi^d_{\boldsymbol{\phi}}}}_{\text{Subpath merging}}\right)=O\left(2\frac{\left(v(G)\right)^2}{K}+4K^2+K^{\frac{3}{2}}\right),
\end{equation*}

Through these calculations, it is evident that the CARSS algorithm, which involves the collaborative efforts of multiple agents to solve TSP as opposed to the conventional algorithm with $K=1$, significantly reduces the temporal and spatial complexities during both training and testing phases. Specifically, the CARSS algorithm achieves a complexity reduction of approximately $\frac{1}{K}$ times that of the original algorithm. This reduction in complexity translates to a substantial enhancement in the scalability of the model using the same computational resources.

\section{Experiments}

In this section, we outline the training process and experimental results of the CARSS algorithm. The training and test datasets are prepared in alignment with \citet{Kool2018AttentionLT}. All instance vertices are drawn from a uniform distribution $U_{[0,1]\times[0,1]}$. The instance sizes, $v(G)$, are set to $\{100, 200, 500, 1000\}$, and the number of agents, $K$, ranges from $\{2,3,\ldots,10, 20, 25\}$. Both training and testing are performed on a GeForce RTX 3090 GPU, where instances with sizes less than 100 utilize a single GPU, while the rest employ two GPUs; however, testing is executed on a single GPU. The decoding strategy involves a greedy approach, selecting the action with the highest probability from the model's action distribution. The optimization gap is computed as $(\text{Obj.} / \text{BKS} - 1) \times 100\%$, with $\text{Obj.}$ representing the cost associated with a solution calculated by a specific algorithm, and $\text{BKS}$ denoting the cost of the instance's optimal solution.

The model's hyperparameters are largely consistent with \citet{Kool2018AttentionLT}. The vertex parameters and hidden layer dimensions of the feedforward neural networks are set to $d_v=256$ and $d_f=512$ respectively. Within the generated subpath model, $H=8$ attention heads are employed. The encoder comprises $L^{\text{enc}_v}=3$ layers of vertex feature aggregation attention, $L^{\text{enc}_a}=3$ layers of agent feature aggregation attention, and the decoder has a single attention layer $L^{\text{dec}_g}=1$. Here, the superscripts $\text{enc}_v$, $\text{enc}_a$, and $\text{dec}_g$ correspond to the encoder for vertex features, encoder for agent features, and decoder for generating policies, respectively. In the subpath merging model, $H=8$ attention heads are used, and in both the encoder and decoder, there are $L^{\text{enc}_{v'}}=3$ and $L^{\text{dec}_c}=1$ attention layers, where the superscripts $\text{enc}_{v'}$ and $\text{dec}_c$ pertain to the vertex encoder and policy decoder, respectively. The model undergoes $E=100$ iterations, with each iteration comprising $B=1000$ batches, and each batch containing 512 instances. The learning rate remains fixed at $10^{-4}$.

Under the aforementioned settings, training a single iteration of this model on a GeForce RTX 3090 GPU takes approximately 10 to 25 minutes for instances with 100 vertices, around 25 to 31 minutes for instances with 200 vertices, and about 49 minutes for instances with 500 vertices. It's noteworthy that the training time for models with the same training set size varies based on the number of agents; larger numbers of agents correspond to shorter training times. For the single-agent Attention Model (AM) proposed by \citet{Kool2018AttentionLT}, its training times on smaller instances align with those of the CARSS algorithm, potentially due to the relatively high number of sequential execution steps during training or excessive sampling, which suggests room for optimization. As for memory consumption, the model can be trained on instances with 100 or 200 vertices using a single GPU, consuming up to a maximum of 12000 MiB of memory. However, for instances with 500 vertices, two GPUs are required, with each consuming around 16000 MiB of memory. On the other hand, the AM model requires two GPUs for training on instances with 200 vertices, consuming approximately 15000 MiB of memory per card. For larger instances, dual-GPU training is infeasible. This highlights the substantial memory optimization improvements achieved by the CARSS algorithm during training.

\subsection{Performance on Random Instances}

As shown in \ref{tab:carss_random}, we employed the CARSS algorithm to conduct tests on randomly generated instances with a maximum size of 1000 vertices. The Best Known Solution (BKS) was obtained using solvers such as Concorde or Gurobi. Since conventional reinforcement learning-based solving methods perform worse than 2-opt and insertion algorithms for instances of this size, only these few algorithms were included in the comparison. Test instances were generated randomly within the domain $U_{[0,1]\times[0,1]}$, with each instance type consisting of $10000/v(G)$ samples. During testing, 4096 results were obtained using greedy decoding for each instance, with the best result and solving time reported. The average values were then computed based on instance sets of equivalent scale. In the "Solver" column, "AM (sample)" represents the sample decoding version of the single-agent algorithm proposed by Kool et al. \citep{Kool2018AttentionLT}. On the other hand, $\text{CARSS}(v(G),K)$ indicates the CARSS algorithm trained with $K$ agents on a graph of size $v(G)$. As Gurobi's solving time becomes prohibitively long for larger instances, it wasn't employed to solve instances with 500 and 1000 vertices, and thus, "–" is used to indicate untested results.

Observing the results, it is evident that for instances with 100 vertices, CARSS (100,2) outperforms the farthest insertion algorithm but slightly lags behind AM (sample). For instances with 200 vertices, the performance of CARSS (100,4) is on par with the farthest insertion method, and superior to AM (sample). As the instance size increases to 500 or 1000 vertices, the optimization gap of CARSS (500,20) is inferior to the nearest insertion algorithm but far better than AM (sample). Even with the increase in sampling iterations, the algorithm retains the potential to achieve better solutions. In terms of testing time, the CARSS algorithm consistently outperforms AM (sample).

\begin{table}[!htbp]
\centering
\caption{Results of CARSS algorithm on random instances}
\label{tab:carss_random}
\begin{adjustbox}{center}
\begin{tabular}{{l|rrr|rrr|rrr|rrr}}
\toprule
Problem Size & \multicolumn{3}{c}{100} & \multicolumn{3}{c}{200} & \multicolumn{3}{c}{500} & \multicolumn{3}{c}{1000} \\
Algorithm & Obj. & Gap & Time & Obj. & Gap & Time & Obj. & Gap & Time & Obj. & Gap & Time \\
\midrule
Concorde & 7.74 & 0.00\% & 0.189s & 10.71 & 0.00\% & 1.015s & 16.55 & 0.00\% & 18.844s & 23.09 & 0.00\% & 1.366m \\
Gurobi & 7.74 & 0.00\% & 1.008s & 10.71 & 0.00\% & 14.585s &  & - &  &  & - &  \\
2-opt & 8.34 & 7.79\% & 0.198s & 11.67 & 8.94\% & 0.606s & 18.20 & 9.98\% & 2.948s & 25.60 & 10.90\% & 31.792s \\
FI & 8.34 & 7.85\% & 0.006s & 11.68 & 9.06\% & 0.022s & 18.26 & 10.37\% & 0.160s & 25.74 & 11.52\% & 1.014s \\
RI & 8.51 & 9.95\% & 0.004s & 11.94 & 11.54\% & 0.009s & 18.46 & 11.56\% & 0.038s & 26.10 & 13.07\% & 0.111s \\
NI & 9.45 & 22.20\% & 0.006s & 13.28 & 23.97\% & 0.022s & 20.63 & 24.66\% & 0.153s & 28.93 & 25.32\% & 0.999s \\
\midrule
AM (sample) & \textbf{7.92} & \textbf{2.39\%} & 1.119m & \textbf{11.50} & \textbf{7.48\%} & 1.547m & 22.65 & 36.82\% & 3.180m & 42.94 & 85.96\% & 6.482m \\
CARSS (100, 2) & 8.09 & 4.53\% & 7.998s & 12.11 & 13.03\% & 16.577s & 21.87 & 32.15\% & 44.526s & 35.28 & 52.83\% & 1.671m \\
CARSS (100, 3) & 8.15 & 5.39\% & 6.631s & 12.13 & 13.25\% & 12.985s & 21.78 & 31.63\% & 34.216s & 34.97 & 51.48\% & 1.274m \\
CARSS (100, 4) & 8.12 & 4.93\% & 5.589s & 12.00 & 12.02\% & 11.232s & 21.24 & 28.36\% & 29.356s & 33.85 & 46.62\% & 1.081m \\
CARSS (100, 5) & 8.15 & 5.34\% & 5.372s & 12.03 & 12.36\% & 10.323s & 21.17 & 27.93\% & 26.184s & 33.32 & 44.36\% & 58.039s \\
CARSS (100, 6) & 8.23 & 6.44\% & 5.239s & 12.26 & 14.51\% & 9.595s & 21.56 & 30.30\% & 24.265s & 34.03 & 47.40\% & 53.644s \\
CARSS (100, 7) & 8.34 & 7.87\% & 4.771s & 12.33 & 15.10\% & 9.389s & 21.71 & 31.22\% & 22.947s & 34.14 & 47.90\% & 50.660s \\
CARSS (100, 8) & 8.34 & 7.83\% & 5.345s & 12.33 & 15.10\% & 10.649s & 22.04 & 33.20\% & 23.163s & 34.78 & 50.67\% & 50.139s \\
CARSS (100, 9) & 8.56 & 10.60\% & 4.827s & 12.70 & 18.57\% & 9.012s & 22.47 & 35.77\% & 22.853s & 36.28 & 57.16\% & 46.253s \\
CARSS (100, 10) & 8.18 & 5.76\% & 7.802s & 12.13 & 13.23\% & 11.628s & 21.37 & 29.12\% & 23.896s & 34.08 & 47.60\% & 49.034s \\
CARSS (200, 5) & 8.23 & 6.36\% & 5.529s & 12.03 & 12.35\% & 10.327s & 20.81 & 25.76\% & 26.170s & 32.56 & 41.04\% & 57.606s \\
CARSS (200, 10) & 8.23 & 6.38\% & 7.801s & 12.10 & 12.98\% & 11.667s & 20.84 & 25.95\% & 24.418s & 32.29 & 39.88\% & 49.065s \\
CARSS (500, 20) & 8.13 & 5.08\% & 33.401s & 12.17 & 13.61\% & 36.003s & \textbf{20.58} & \textbf{24.35\%} & 46.409s & \textbf{31.03} & \textbf{34.41\%} & 1.098m \\
\bottomrule
\end{tabular}
\end{adjustbox}  % [How can I center a too wide table?](https://tex.stackexchange.com/a/39436)
\end{table}

\subsection{Sensitivity Analysis}

\begin{figure}[!htbp]
	\centering
	\begin{subfigure}{0.8\textwidth}
		\includegraphics[width=\textwidth]{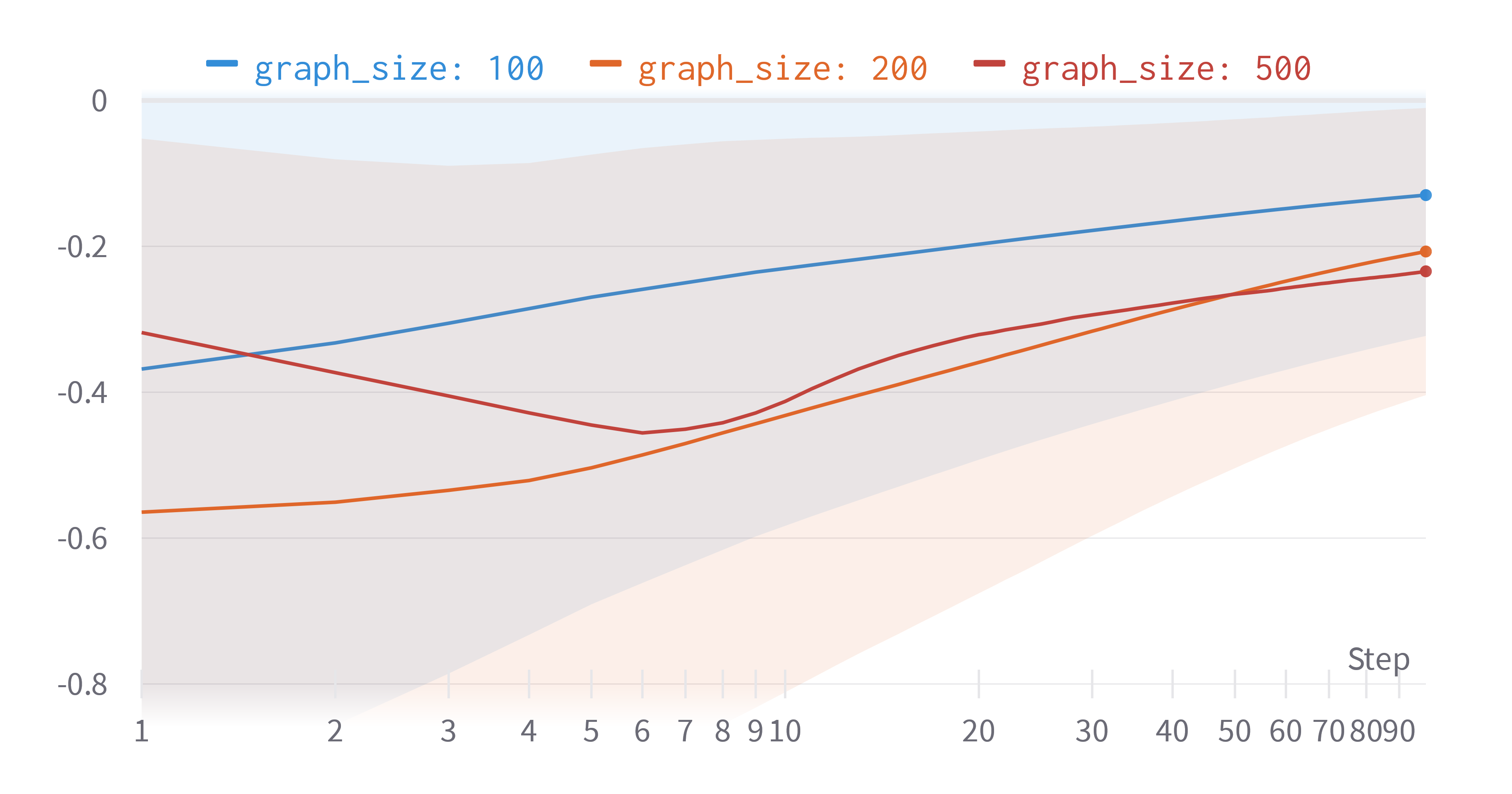}
	\end{subfigure}
    \vskip\baselineskip
	\begin{subfigure}{0.8\textwidth}
		\includegraphics[width=\textwidth]{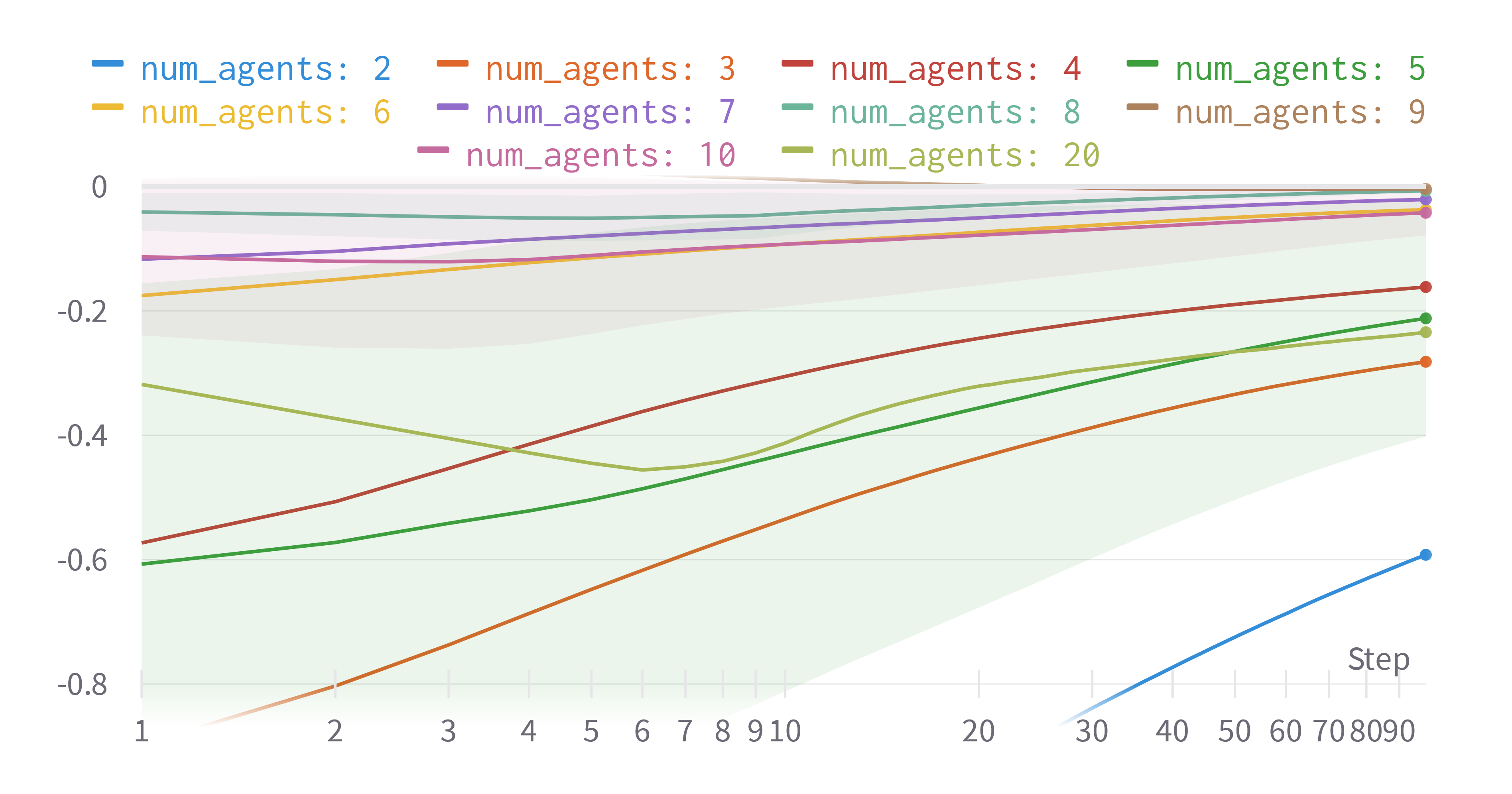}
	\end{subfigure}
    \vskip\baselineskip
	\begin{subfigure}{0.8\textwidth}
		\includegraphics[width=\textwidth]{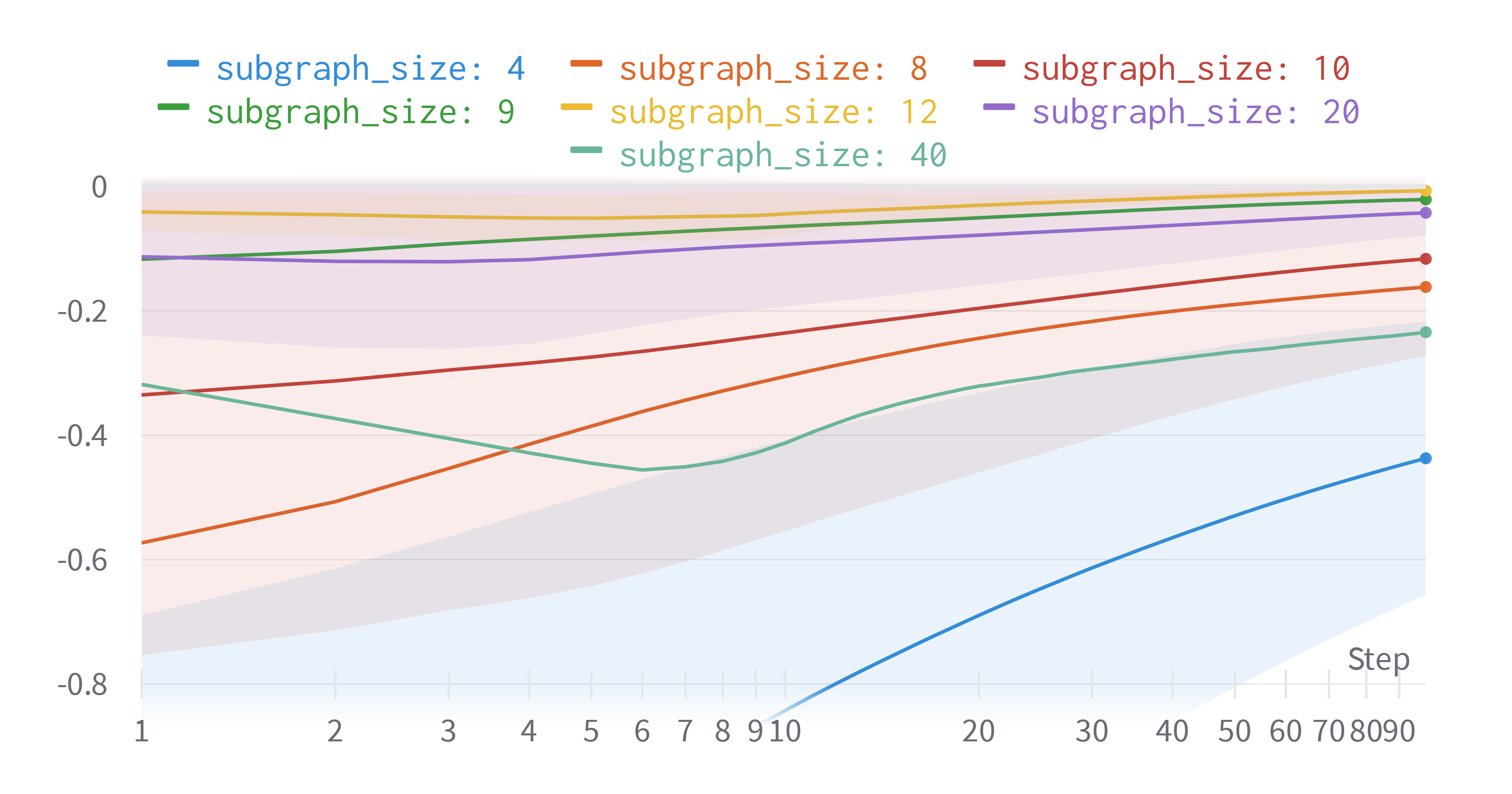}
	\end{subfigure}
	\caption{Variation of training loss with number of iterations of CARSS algorithm}
	\label{fig:carss_loss}
\end{figure}

This segment delves into the relationship between the training loss of the subpath generation model and the problem's scale and the number of agents involved. The training process is categorized into three groups based on the instance size $v(G)$, the number of agents $K$, and the subproblem scale in the subpath merging phase $(K+|I|)$. Figure \ref{fig:carss_loss} illustrates the relationship between the training loss $L(\boldsymbol{\theta})$ and the number of iterations $E$. Solid lines represent the mean of the loss within each category, while shaded regions depict the fluctuation range in terms of standard deviation. A higher absolute value of the loss implies a greater potential for improvement with more training iterations. 

From the graph, we observe that the training loss of the subpath generation model exhibits similar trends across various problem scales. This suggests the stability of the CARSS algorithm's performance when training on different problem sizes, without encountering training difficulties due to excessively large problem scales. However, as the number of agents or the subproblem scale in the subpath merging phase increases, the absolute value of the loss diminishes and its rate of reduction slows down over the training process. This phenomenon can be attributed to the multi-modal nature of the cooperative multi-agent environment.

\subsection{Example Solutions}

\begin{figure}[!htbp]
	\centering
	\begin{subfigure}{.3\textwidth}
		\includegraphics[width=\textwidth]{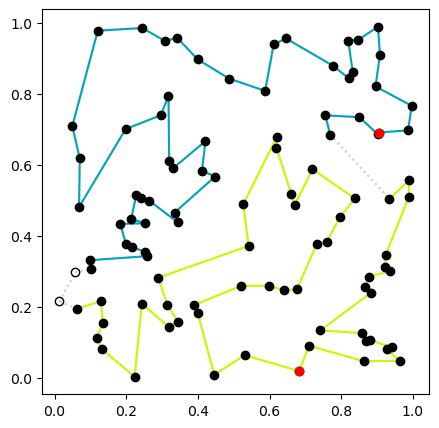}
		\caption{2 agents}
		\label{subfig:marltsp_na2}
	\end{subfigure}
	\begin{subfigure}{.3\textwidth}
		\includegraphics[width=\textwidth]{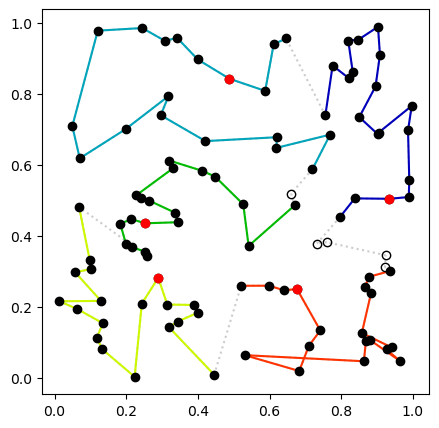}
		\caption{5 agents}
		\label{subfig:marltsp_na5}
	\end{subfigure}
	\begin{subfigure}{.3\textwidth}
		\includegraphics[width=\textwidth]{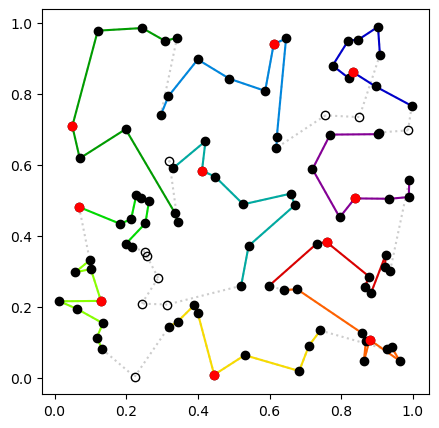}
		\caption{10 agents}
		\label{subfig:marltsp_na10}
	\end{subfigure}
	\caption{Example of CARSS algorithm's solution on a graph with 100 vertices}
	\label{fig:carss_solution}
\end{figure}

As depicted in Figure \ref{fig:carss_solution}, we selected an instance with a size of 100 and employed the CARSS algorithm with three distinct configurations, where the number of agents $K$ was set to $\{2, 5, 10\}$, to solve it using a greedy decoding strategy. In the figure, solid black dots represent vertices in the instance, red solid dots indicate the initial vertices chosen by each agent, and hollow black dots depict isolated vertices not selected by any agent, totaling $|I|$ in number. Different colored solid lines correspond to subpaths of different agents in the final solution, amounting to a total of $K$ subpaths, while dashed lines symbolize the $K+|I|$ edges added to connect all agent subpaths and isolated vertices into a cycle during the subpath merging phase. From the illustration, it becomes apparent that the CARSS algorithm adeptly captures the characteristics of optimal solutions for TSP. Solutions exhibit superior quality with fewer agents, and the traveling salesman's route demonstrates minimal instances of overlapping. However, as the number of agents increases, a potential decline in algorithm performance can be observed. This might arise from the simplicity in the mapping of the vertex index $U^{n,K}_{t}$, output by the policy network during the subpath generation phase, to the action $A^{n,k}_{t}$ at time $t$. In this process, a choice is made to connect the chosen vertex to the nearest end of the subpath, rather than "inserting" the selected vertex into the current subpath, as seen in algorithms like the farthest insertion method. This discrepancy could lead to a decline in performance. This issue is evident in the solution provided by CARSS (100,10) for the loop in the upper right corner of the route, where the corresponding subpath formed by the blue agent shows such behavior.

\section{Conclusion}

In this paper, we introduced CARSS algorithm, a groundbreaking approach for solving TSP using cooperative MARL. CARSS strategically decomposes the TSP solving process into subpath generation and subpath merging steps, leveraging the power of cooperative MARL to tackle the challenges posed by large-scale instances. By employing attention mechanisms for feature embedding and parameterization, CARSS enhances the agents' ability to learn and generate high-quality solutions. The independent REINFORCE algorithm facilitates the training of the CARSS model, contributing to its efficiency and effectiveness.

Our contributions to the field are threefold: firstly, the introduction of CARSS, an innovative algorithmic framework that harnesses cooperative MARL for TSP solving; secondly, the integration of attention mechanisms, which significantly elevate the agents' learning capabilities; and thirdly, the empirical demonstration of CARSS's superiority over single-agent alternatives. Through comprehensive experiments, we showed that CARSS outperforms conventional approaches in terms of delivering reduced memory consumption, improved scalability, and notable reductions in testing time and optimization gaps for large-scale TSP instances.

As the field of combinatorial optimization and reinforcement learning continues to evolve, CARSS presents a robust strategy that capitalizes on the synergy of multiple agents and attention mechanisms. While our work demonstrates remarkable advancements in tackling TSP, future research could explore the application of CARSS to other combinatorial optimization problems and delve deeper into optimizing the attention mechanisms to further enhance the agents' learning efficiency. We anticipate that CARSS will play a pivotal role in advancing the capabilities of MARL in addressing complex real-world optimization challenges.

In conclusion, our study underscores the effectiveness of the CARSS algorithm, shedding light on its potential to revolutionize the way we approach TSP and related problems. By combining the strengths of cooperative MARL, attention mechanisms, and subpath synthesis, CARSS represents a significant stride toward efficient and scalable solutions for the TSP.

\section*{Acknowledgments}

This research was supported by National Key R\&D Program of China (2021YFA1000403), the National Natural Science Foundation of China (Nos. 11991022), the Strategic Priority Research Program of Chinese Academy of Sciences (Grant No. XDA27000000) and the Fundamental Research Funds for the Central Universities.

\bibliographystyle{unsrtnat}

\end{document}